\documentclass[preprint,12pt]{elsarticle}

\usepackage{amssymb}
\usepackage{amsmath}
\usepackage{booktabs}
\usepackage{multirow}
\usepackage{array}
\usepackage{hyperref}
\usepackage{lineno}
\usepackage{graphicx}
\usepackage{makecell}
\graphicspath{{figures/}}
\usepackage{xurl}

\usepackage{xcolor}

\journal{Biomedical Signal Processing and Control}

\begin{document}

\begin{frontmatter}

\title{Spectral Features Dominate BCG Respiratory-Event Detection: A Large-Scale Patient-Independent Comparison of Feature Groups in Sleep Apnea Patients}

\author[rug]{Israel Campero Jurado\corref{cor1}}
\ead{i.c.j.campero.jurado@rug.nl}

\author[usz]{Zoe Bousraou}

\author[usz]{Lara Benning}

\author[eth]{Sara Padilla Neira}

\author[eth]{Alexander Breuss}

\author[eth]{Robert Riener}

\author[usz]{Esther Irene Schwarz}

\author[rug]{Elisabeth Wilhelm}

\cortext[cor1]{Corresponding author}

\affiliation[rug]{
  organization={Department of  Engineering and Technology Institute Groningen, Faculty of Science and Engineering},
  addressline={University of Groningen},
  %city={Groningen},
  %postcode={9747 AG},
  country={The Netherlands}
}

\affiliation[usz]{
  organization={Department of Pulmonology},
  addressline={University Hospital Zürich},
  %city={Zürich},
  %postcode={9747 AG},
  country={Switzerland}
}

\affiliation[eth]{
  organization={Sensory-Motor Systems Lab},
  addressline={ETH Zürich},
  %city={Zürich},
  %postcode={9747 AG},
  country={Switzerland}
}
%% ────────────────────────────────────────────────────────────────────────────

\begin{abstract} 

Unobtrusive ballistocardiographic~(BCG) sensing is a promising modality for long-term sleep-apnea monitoring, yet it remains unclear which signal features are most discriminative for respiratory-event detection. We present a literature-guided, patient-independent comparison of ten BCG feature groups using a 512-sensor capacitive pressure mat recorded simultaneously with respiratory polygraphy in 155 patients~(52 female, 103 male) undergoing an in-laboratory cardiorespiratory polygraphy for obstructive sleep apnea. Features were extracted from six spatially distinct signal channels, yielding a 191-dimensional feature vector spanning general statistical, time-domain, frequency-domain, wavelet, frame-energy, and nonlinear complexity descriptors.

Under strict leave-one-patient-out cross-validation for binary classification of respiratory-event windows versus event-free reference windows, Random Forest and Histogram Gradient Boosting achieved AUC-ROC of $0.967$ and $0.969$ and AUC-PR of $0.977$ and $0.979$, respectively. Feature-importance analysis revealed that frequency-domain features dominate discrimination: breathing-band power in the $0.1$--$0.4$\,Hz range accounted for 30.3\% of total discriminative information across all spatial channels, and  Fast Fourier Transform (FFT) spectral-shape descriptors of the adaptively preprocessed channel contributed a further 15.1\%. Area Under the Curve (AUC) and curve-length features provided the main complementary time-domain evidence~(21.5\%), whereas wavelet-derived and nonlinear features contributed smaller secondary effects~(10.4\% combined across 59 features). Frequency-domain and time-domain features together accounted for 67\% of total discriminative information, demonstrating that a compact, interpretable subset of the full feature library achieves clinically relevant performance under patient-independent validation and providing an empirical basis for feature selection in future BCG systems.

\end{abstract}

\begin{keyword}
Obstructive sleep apnea \sep BCG \sep Unobtrusive sensing \sep
Feature extraction \sep Machine learning \sep Polygraphy synchronisation \sep
Signal processing
\end{keyword}

\end{frontmatter}

%% ============================================================================

%% ============================================================================
\section{Introduction}
\label{sec:intro}
%% ============================================================================
Ballistocardiographic (BCG) sensing using force or pressure sensors integrated into or placed beneath the mattress, in the bed frame, or at the foot of the bed is an attractive modality for long-term sleep monitoring because it captures respiratory effort, posture, and gross body movement without requiring sensors to be attached to the body, and is therefore comfortable enough for repeated overnight use~\cite{vitazkova2024advances, Inan2015BallistocardiographyAS,bruser2012multi}. Bed-based BCG has been investigated for sleep-disordered breathing~(SDB) detection in a number of studies, and it has been shown that the BCG signal carries information sufficient to distinguish respiratory events from normal-breathing segments~\cite{waltisberg2016detecting,wang2017assessing, zhao2015identifying,gao2019obstructive,sadek2020new,qi2023mattress}. However, these studies differ substantially in sensing hardware, feature definitions, prediction targets, and validation protocols~(see Table~\ref{tab:prior_features}). As a direct consequence, it remains unclear which feature groups contribute most to the discrimination of respiratory-event windows from normal-breathing windows in BCG-based detection. Prior work has used general statistical descriptors~\cite{waltisberg2016detecting,wang2017assessing, gao2019obstructive,sadek2020new,qi2023mattress}, time-domain waveform features~\cite{waltisberg2016detecting,zhao2015identifying,qi2023mattress}, frequency-domain band power~\cite{waltisberg2016detecting,wang2017assessing, gao2019obstructive}, FFT spectral-shape descriptors~\cite{mendez2010automatic,dafna2013automatic}, wavelet-derived features~\cite{zhao2015identifying,wang2017assessing, qi2023mattress}, and nonlinear complexity measures~\cite{zhao2015identifying,wang2017assessing,mendez2010automatic}, but none has compared all of these under a single, patient-independent experimental protocol.

Establishing a feature ranking matters for two practical reasons. First, it identifies which BCG signal properties are indicative of apneic or hypopneic events, grounding the classification in signal physiology rather than in benchmark results alone. Second, it provides an empirical basis for deciding which feature groups are worth computing in future real-time or resource-constrained systems and which can be omitted without meaningful performance loss. The goal of the present work is therefore to answer the following research questions under strict leave-one-patient-out~(LOPO) cross-validation in a cohort of 155 patients:

\begin{enumerate}
  \item Which feature categories carry the most discriminative information
        for BCG respiratory-event detection?
  \item Does adaptive preprocessing (high-pass filtering, quality-score-based
        sensor selection, and polarity-aligned averaging) amplify feature
        discriminability compared with raw spatial aggregates?
  \item Which feature categories provide complementary discriminative
        information beyond the dominant tier?
\end{enumerate}
We extracted 191 features organised into ten groups, informed by a structured review of prior BCG, electrocardiogram (ECG), and acoustic SDB studies, and applied them to six signal channels derived from a 512-sensor capacitive BCG recorded synchronously with respiratory polygraphy. Random Forest feature importance~(mean decrease in impurity) provides the group ranking. We address these questions through a systematic study in which all ten feature groups are applied to six signal channels derived from a 512-sensor capacitive BCG recorded synchronously with respiratory polygraphy in 155 patients, and evaluated under strict leave-one-patient-out cross-validation. Their discriminative contributions are quantified using Random Forest mean decrease in impurity, providing a standardised, patient-independent feature-group ranking for BCG respiratory-event detection. %The principal finding is that frequency-domain features~(spectral-shape descriptors and breathing-band power) jointly account for 45\% of the total discriminative information, time-domain AUC and curve-length features account for a further 22\%, and all remaining groups~(wavelet, statistical, nonlinear) together contribute 33\% distributed across 143 features.

%% ============================================================================
%% ============================================================================
\section{Background and Related Work} \label{sec:background}

\subsection{BCG Sensing for Sleep-Disordered Breathing} \label{subsec:bcg_sensing}

BCG sensing measures small changes of mechanical forces and body motions generated by cardiovascular activity, especially the inertial effects of cardiac ejection and blood flow~\cite{Sadek2019BallistocardiogramSP, Balali2022InvestigatingCI}. Bed-based sensors can also capture respiratory effort, posture, and gross body movement~\cite{giovangrandi2011ballistocardiography}. When a sensor is placed beneath the mattress or bed sheet, the measured signal is not a pure physiological signal, but a superposition of several mechanical sources. Slow components are mainly associated with body load, posture, and gradual position changes~\cite{nickerson1950diagnosis,kim2016ballistocardiogram,paalasmaa2012unobtrusive}. Respiratory effort appears as a low-frequency oscillatory component. Cardiac micromotion related to blood ejection from the heart and blood flow through the aorta contributes additional low-frequency and higher-frequency components to the BCG signal. The relative contribution of these components depends on sensor type, mattress coupling, body posture, and preprocessing~\cite{gomez2014towards,albukhari2019bed}.

This mixed mechanical origin makes BCG sensing attractive but also challenging for sleep-disordered-breathing detection. During respiratory events, changes in thoracoabdominal expansion, inverse or paradoxical breathing patterns, recovery breaths, and arousal-related changes in heart rate can alter the frequency content and waveform morphology of the recorded signal~\cite{Pysick2026SevereNR, Mazzotti2018OpportunitiesFU}. Careful feature engineering is therefore important because appropriate features may capture subtle event-related changes in respiratory modulation, cardiac-related dynamics, and overall signal shape, thereby improving classifier performance. 

\subsection{Feature-Based Approaches in Prior Bed-Based OSA Work} \label{subsec:prior_features}

Table~\ref{tab:prior_features} summarises representative BCG and bed-pressure studies for sleep-disordered-breathing detection. Related ECG and acoustic work is discussed where it has informed the selection of features included in the present study, since these modalities target the same underlying physiological events and their feature choices provide useful guidance for pressure-mat feature design.

\begin{table*}[h!]
\caption{Feature categories used in prior BCG and bed-pressure studies for
sleep-disordered breathing or OSA-related detection. \checkmark~=~used as a
classifier or decision feature; preproc.=used during preprocessing, signal
extraction, or fusion but not as a final classifier-feature category;
NR~=not reported. PVDF stands for polyvinylidene fluoride, a piezoelectric polymer material.}
\label{tab:prior_features}
\centering
\scriptsize
\setlength{\tabcolsep}{3.0pt}
\resizebox{\textwidth}{!}{%
\begin{tabular}{llcccccc}
\toprule
\textbf{Study} & \textbf{Signal / modality} &
\makecell{\textbf{General}\\\textbf{stat.}} &
\makecell{\textbf{Time}\\\textbf{domain}} &
\makecell{\textbf{Frequency}\\\textbf{domain}} &
\makecell{\textbf{Wavelet}\\\textbf{/ TF}} &
\makecell{\textbf{Nonlinear}\\\textbf{/ complexity}} &
\makecell{\textbf{Spatial}\\\textbf{/ fusion}} \\
\midrule
Zhao et al.~\cite{zhao2015identifying}
  & BCG-derived HBI + STC-Min breathing effort
  & NR & \checkmark & NR & preproc. & \checkmark & NR \\

Waltisberg et al.~\cite{waltisberg2016detecting}
  & 8-sensor under-mattress strain-gauge array
  & \checkmark & \checkmark & \checkmark & NR & NR & \checkmark \\

Wang et al.~\cite{wang2017assessing}
  & BCG-derived RR intervals
  & \checkmark & NR & \checkmark & preproc. & \checkmark & NR \\

Gao et al.~\cite{gao2019obstructive}
  & Mattress BCG / respiratory signal; HRV + similarity features
  & \checkmark & NR & \checkmark & NR & NR & NR \\

Sadek et al.~\cite{sadek2020new}
  & Under-mattress microbend fiber-optic sensor
  & \checkmark & NR & NR & preproc. & NR & NR \\

Qi et al.~\cite{qi2023mattress}
  & PVDF mattress pressure sensor; respiratory waveform similarity
  & \checkmark & \checkmark & NR & preproc. & NR & NR \\
\bottomrule
\end{tabular}
}
\end{table*}

As Table~\ref{tab:prior_features} shows, no single study has used all feature categories simultaneously, and no study has compared their relative discriminative contribution under patient-independent validation. General statistical features summarising central tendency and variability~\cite{wang2017assessing,gao2019obstructive,sadek2020new, qi2023mattress}, time-domain waveform descriptors~\cite{zhao2015identifying,waltisberg2016detecting, qi2023mattress}, and frequency-domain band-power features~\cite{waltisberg2016detecting,wang2017assessing,gao2019obstructive} are the most consistently used categories. Wavelet-based methods and nonlinear complexity measures~(sample entropy, detrended fluctuation analysis) have also been applied, but predominantly for signal separation or cardiac-interval analysis rather than as direct classifier inputs~\cite{zhao2015identifying,wang2017assessing,qi2023mattress}. This heterogeneity motivates the systematic comparison presented here.

\subsection{Spatial Information in Bed-Based Sensing} \label{subsec:spatial}

Many BCG systems rely on a single sensor or a global aggregate, which provides only a limited projection of the spatially distributed mechanical activity~\cite{paalasmaa2012unobtrusive,sadek2020new, qi2023mattress}. High-resolution pressure mats may provide additional information because thoracoabdominal motion, posture-dependent coupling, and asynchronous or inverse breathing patterns can have different expressions across the upper, middle, and lower body regions. Prior work has shown that multi-sensor fusion can improve apnea and movement detection~\cite{waltisberg2016detecting,bruser2012multi, hu2025chmmconvscalenet}, but whether discriminative information is best captured by a global signal, regional aggregates, or adaptively selected local sensors remains unclear. This study evaluates six spatial signal channels to address this question directly.

%============================================================================ 
\section{Materials and Methods} \label{sec:methods} %% ============================================================================

\subsection{Participants and Data Collection} \label{subsec:participants}

\begin{figure}[ht]
  \centering
  \includegraphics[width=\textwidth]{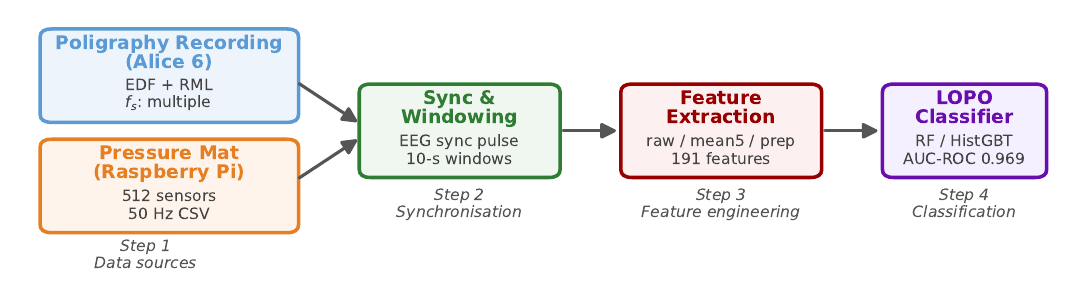}
  \caption{End-to-end processing pipeline. \textbf{Step~1:} simultaneous acquisition from the Alice~6 PSG system and a Raspberry Pi data logger recording the
  $16{\times}32$ BCG at 50\,Hz. \textbf{Step~2:} EEG sync-pulse alignment maps expert-scored events onto the BCG timeline. \textbf{Step~3:} six signal channels are derived and processed through standardised feature groups, yielding a 191-dimensional feature vector per window. \textbf{Step~4:} leave-one-patient-out cross-validated classifiers produce binary predictions~(respiratory event vs.\ event-free window).}
  \label{fig:pipeline}
\end{figure}

Data were recorded at the Sleep Disorders Centre of the Department of Pulmonology, University Hospital Zürich~(USZ). This clinical trial was approved by the Cantonal Ethics Committee of Zurich (Ethikkommission Zürich), Switzerland (BASEC No. 2025-00339). The study was conducted in accordance with the principles of the Declaration of Helsinki, the Swiss Human Research Act (HRA), and Good Clinical Practice (GCP) guidelines. Written informed consent was obtained from all participants prior to enrolment. A custom sensor unit consisting of a Raspberry Pi~(RPi) data logger and a $16 \times 32$ capacitive BCG~(512 sensors, $f_s = 50$\,Hz) placed beneath the bed sheet was deployed alongside the standard hospital polygraphy workflow using the Philips Respironics Alice~6 system~\cite{meszaros2025robotic}. The mat was placed within the mattress, between the cushioning layer and the mattress cover, over which the sheet was then placed. Expert scoring of polygraphy recordings was performed according to AASM~2020 rules~\cite{berry2012rules, ramar2021sleep}. The overall procedure described in this section is summarised in Figure~\ref{fig:pipeline}.

\subsection{Polygraphy--BCG Synchronisation} \label{subsec:sync}

Signals from the polygraphy system and the Raspberry Pi BCG logger were aligned using a dedicated EEG sync-pulse protocol. The two systems operate on independent clocks, so synchronisation was required before polygraphy scored respiratory events could be mapped onto the BCG timeline.

The synchronisation protocol consisted of two phases. In the first phase (duration approximately 10\,s), a pseudo-random binary sequence~(PRBS) was coupled into one EEG channel of the polygraphy system as a large-amplitude positive square wave. The Raspberry Pi generated this digital sequence, stored it in the Pi sync CSV file together with RPi timestamps, and used it to drive a GPIO pin connected to the EEG input of the polygraphy device. In the second phase, 18-second frames were transmitted at 2\,Hz. Each frame encoded a 16-bit timestamp using on--off keyed binary encoding: a HIGH--LOW start token followed by 16 one-second bit slots, with two samples per slot and 36 samples per frame. The bit value was determined by the state of the first sample in each slot, where HIGH represented 1 and LOW represented 0.

The polygraphy EEG sync channel was binarised by thresholding at the 90th percentile of the raw channel amplitude. The RPi and polygraphy sync streams were then decoded independently by searching for valid start tokens, verifying inter-sample timing consistency, and decoding the transmitted 16-bit values. The synchronisation offset was computed from the first matched timestamp sequence: 

\begin{equation} 
\Delta t = t_{\mathrm{Pi}}^{(q)} - t_{\mathrm{polygraphy}}^{(q)}, 
\label{eq:sync_offset} 
\end{equation} 

where $q$ denotes the first timestamp sequence decoded in both systems, $t_{\mathrm{Pi}}^{(q)}$ is the RPi timestamp of that sequence, and $t_{\mathrm{polygraphy}}^{(q)}$ is the corresponding polygraphy timestamp.

Expert-scored redline markup language (RML) respiratory events were then mapped to the BCG timeline. The respiratory-event labels included Obstructive Apnea, Hypopnea, Obstructive Hypopnea, Central Apnea, Mixed Apnea, and Central Hypopnea. For an event with polygraphy start time $t_{\mathrm{polygraphy},e}$, the corresponding RPi time was computed as 

\begin{equation} 
t_{\mathrm{Pi},e} = t_{\mathrm{polygraphy},e} + \Delta t, 
\label{eq:event_time_mapping} 
\end{equation} 

where $t_{\mathrm{Pi},e}$ is the event start time on the RPi clock. The BCG sample index was then obtained as 

\begin{equation}   
n_e =   \left\lfloor   \left(t_{\mathrm{Pi},e}-t_{\mathrm{Pi},0}\right) f_s   \right\rceil,   
\label{eq:event_sample_mapping} 
\end{equation} 

where $n_e$ is the mapped BCG sample index, $t_{\mathrm{Pi},0}$ is the RPi timestamp of the first BCG sample, $f_s=50$\,Hz is the BCG sampling frequency, and $\lfloor \cdot \rceil$ denotes rounding to the nearest integer.

\subsection{Signal Channels and Window Selection} \label{subsec:channels_windows}

Feature extraction was performed on one-dimensional signals derived from the 512-channel BCG. The purpose of this step was to obtain comparable spatial representations before computing the same feature groups. Six signal channels were used: four raw spatial aggregates, one raw top-sensor average, and one adaptively preprocessed top-sensor average.

\paragraph{Raw spatial aggregates} The channel \texttt{raw\_total} was defined as the pointwise sum of all 512 raw BCG sensor values (16 $\times$ 32 sensors grid). This channel summarises the total pressure variation measured by the mat and is sensitive to posture, gross movement, and large-scale thoracoabdominal pressure modulation. Three additional zonal aggregates were computed by partitioning the mat along the body axis: \texttt{raw\_upper}~(rows~0--3, shoulder/upper-back region), \texttt{raw\_mid}~(rows~4--11, mid-back), and \texttt{raw\_lower}~(rows~12--15, lower-back region). These channels provide coarse anatomical spatial information while preserving raw sensor amplitudes.

Let $x_i[n]$ denote the raw signal of sensor $i$ at sample $n$, with
$i=1,\ldots,512$. For any sensor subset $\mathcal{R}$, the corresponding raw
aggregate was computed as

\begin{equation}
x_{\mathcal{R}}[n]
=
\sum_{i \in \mathcal{R}} x_i[n].
\label{eq:raw_aggregate}
\end{equation}

The global raw aggregate, denoted \texttt{raw\_total}, was obtained by setting
$\mathcal{R}$ equal to the full set of 512 sensors. Regional aggregates were
computed by selecting sensors according to their spatial location on the mat.
Specifically, the mat was divided into upper, middle, and lower body regions,
corresponding to the first four sensor rows, the middle eight sensor rows, and
the last four sensor rows, respectively. These aggregates are denoted
\path{raw_upper}, \path{raw_mid}, and \path{raw_lower}.

\paragraph{A selection of top-quality sensors} To obtain a local pressure signal with strong respiratory periodicity, sensors were ranked using a respiratory-quality score (based on autocorrelation). Before computing this score, each sensor was high-pass filtered using a 4th-order Butterworth filter with cut-off frequency $f_c=0.05$\,Hz. This filter was used to suppress slow baseline drift while retaining respiratory, cardiac, and movement-related dynamics above the cut-off frequency.

For each sensor~$i$, the respiratory-quality score $S_i$~Eq.~\ref{eq:thv})
was adapted from the threshold-value criterion proposed by Padilla Neira et al.~\cite{padilla_reference}, which combines an autocorrelation term and a spectral-concentration term to rank
sensors by the strength and periodicity of their respiratory signal.
\begin{equation}
S_i =
\underbrace{
\max_{\tau \in [2,5]\,\mathrm{s}}
\hat{R}_{ii}(\tau)
}_{\text{autocorrelation term}}
+
\underbrace{
\frac{
\displaystyle
\sum_{f_k \in [f_i^\ast-0.1,\,f_i^\ast+0.1]\,\mathrm{Hz}}
P_i(f_k)
}{
\displaystyle
\sum_{f_k \in [0.1,\,0.5]\,\mathrm{Hz}}
P_i(f_k)
}
}_{\text{spectral-concentration term}}.
\label{eq:thv}
\end{equation}
Here, $\hat{R}_{ii}(\tau)$ is the normalised autocorrelation of the high-pass-filtered signal from sensor $i$ at lag $\tau$, and $P_i(f_k)$ is the power spectral density of sensor $i$ at frequency bin $f_k$. The frequency $f_i^\ast$ denotes the dominant frequency of sensor $i$ within the candidate
breathing-frequency range. The numerator of the
spectral-concentration term sums the spectral power in a 0.2\,Hz-wide band centred on $f_i^\ast$, that is, from $f_i^\ast-0.1$\,Hz to $f_i^\ast+0.1$\,Hz. The denominator sums the total spectral power in the broader respiratory band $[0.1,0.5]$\,Hz. Thus, the autocorrelation term
favours sensors with a clear repeated pattern in the expected breathing-lag range of 2--5\,s, while the spectral-concentration term favours sensors whose respiratory-band power is concentrated near a dominant breathing-related
frequency.

The five sensors with the largest values of $S_i$ were selected for each window. Let $\mathcal{T}$ denote this set of selected top sensors, with $|\mathcal{T}|=5$.

\paragraph{Raw top-sensor average} 
The channel \texttt{raw\_mean5} was computed as the pointwise average of the five selected sensors using the raw, unfiltered signals: 
\begin{equation}   
x_{\mathrm{raw\_mean5}}[n]   =   \frac{1}{5}   \sum_{i \in \mathcal{T}} x_i[n].   
\label{eq:raw_mean5} 
\end{equation} 

This channel preserves the raw pressure signal of the selected sensors, including slow baseline components.

\paragraph{Preprocessed} The channel \path{prep} was designed as a locally coherent respiratory-pressure representation, complementary to the unaligned raw global and regional aggregates. It was computed from the top-ranked sensors selected by the adaptive sensor-selection criterion after high-pass filtering and polarity alignment. Polarity alignment was applied only before forming this adaptive top-sensor average, because local mattress deformation and sensor loading can cause respiratory-related changes to appear with opposite signs across selected sensor locations. The highest-ranked selected sensor was used as the reference, and each other selected sensor was inverted when its first-difference trend was more often opposite than aligned with the reference. The raw global and regional aggregates were retained because they describe net pressure redistribution across the full mat or anatomical regions, rather than a polarity-aligned local respiratory waveform.

Let $\tilde{x}_i[n]$ denote the high-pass-filtered signal from selected sensor $i$, and let $r[n]$ denote the high-pass-filtered signal of the
highest-ranked reference sensor. The difference of a signal is defined as $z[n]$ is $\Delta z[n]=z[n+1]-z[n]$. We use $\#\{\cdot\}$ to denote the number of samples satisfying a condition. A selected sensor was assigned polarity
$a_i \in \{-1,1\}$ according to

\begin{equation}
a_i =
\begin{cases}
-1, &
\#\{n:\Delta r[n]\Delta \tilde{x}_i[n] < 0\}
>
\#\{n:\Delta r[n]\Delta \tilde{x}_i[n] > 0\}, \\
1, & \text{otherwise}.
\end{cases}
\label{eq:polarity_alignment}
\end{equation}

Thus, a sensor was inverted when its local temporal changes were more often opposite than aligned with those of the reference sensor. The adaptively preprocessed signal was then computed as

\begin{equation}
x_{\mathrm{prep}}[n]
=
\frac{1}{|\mathcal{T}|}
\sum_{i \in \mathcal{T}} a_i \tilde{x}_i[n],
\label{eq:prep_signal}
\end{equation}

where $\mathcal{T}$ is the set of selected sensors. In this study,
$|\mathcal{T}|=5$.

Thus, \texttt{prep} is a high-pass filtered, quality-selected, polarity-aligned spatial average of the five sensors with the strongest respiratory periodicity.

\paragraph{Window selection} Positive windows were defined from expert-scored respiratory events mapped to the BCG timeline. The positive class label was $y=1$. Events shorter than 10\,s were symmetrically padded to 10\,s. Negative windows were sampled from recording segments free of any annotated event and were assigned label $y=0$. Negative windows had a minimum duration of 30\,s and were sampled with a 30\,s safety buffer from annotated events to reduce contamination by peri-event arousals or movement. For filtering and preprocessing, a 5\,s context was included on both sides of each window and discarded before feature computation.

\subsection{Implemented Feature Groups} \label{subsec:implemented_features}

Categories were implemented as ten concrete feature groups: \path{StandardStat}, \path{InterQuartile}, \path{ZeroCrossing}, \path{AUC}, \path{PSD}, \path{FFT}, \path{Energy}, \path{Entropy}, \path{Wavelet}, and \path{HiguchiFD}.

The six signal channels represented two complementary types of pressure-mat information. The raw global, regional, and selected-sensor aggregate channels were retained to preserve net pressure redistribution across the mat and across anatomical regions. In contrast, \texttt{prep} was designed as an adaptive respiratory waveform representation by applying high-pass filtering, sensor-quality selection, polarity alignment, and averaging of the selected sensors.

Six feature groups were computed for every signal channel: \texttt{StandardStat}, \texttt{InterQuartile}, \texttt{ZeroCrossing}, \texttt{AUC}, \texttt{PSD}, and \texttt{HiguchiFD}. Together, these groups produced 22 common features per channel, allowing direct comparison of feature groups across global, regional, selected-sensor, and adaptively preprocessed representations.

Four additional feature groups were computed only for \texttt{prep}: \texttt{FFT}, \texttt{Energy}, \texttt{Entropy}, and \texttt{Wavelet}. These groups were restricted to \texttt{prep} because they were intended to describe the respiratory-enhanced waveform after sensor selection, polarity alignment, and baseline-drift removal, rather than the unaligned net pressure redistribution captured by the raw aggregate channels. This design avoids mixing two different signal representations while still allowing the common feature groups to compare information across all channels.

Consequently, the five aggregate channels contributed $5\times22=110$ features, while \texttt{prep} contributed $22+59=81$ features. The complete representation therefore contained $110+81=191$ features per window.

\begin{table}[ht]
\caption{Implemented feature groups and their per-channel application.
\checkmark~=~computed for all six channels; \texttt{prep}=computed only for
the adaptively preprocessed channel.}
\label{tab:features}
\centering
\small
\resizebox{\textwidth}{!}{%
\begin{tabular}{llrr}
\toprule
\textbf{Feature group} & \textbf{Features included} & \textbf{Count}
  & \textbf{Applied to} \\
\midrule
\texttt{StandardStat}
  & Mean, median, standard deviation, variance, kurtosis, skewness, range
  & 7 & \checkmark \\

\texttt{InterQuartile}
  & IQR, interdecile range, upper/lower quartile, upper/lower decile
  & 6 & \checkmark \\

\texttt{ZeroCrossing}
  & Zero-crossing count
  & 1 & \checkmark \\

\texttt{AUC}
  & Signed AUC, absolute AUC, length of curve
  & 3 & \checkmark \\

\texttt{PSD}
  & Total power; band power at 0.1--0.6, 0.1--0.4, 0.4--0.8\,Hz
  & 4 & \checkmark \\

\texttt{HiguchiFD}
  & Higuchi fractal dimension
  & 1 & \checkmark \\

\midrule

\texttt{FFT}
  & Dominant frequency, spectral mean/std/centroid/skewness/entropy
  & 6 & \texttt{prep} \\

\texttt{Energy}
  & Total energy, mean power, RMS, frame-energy mean/std/median/max
  & 7 & \texttt{prep} \\

\texttt{Entropy}
  & Shannon entropy of signal-value histogram
  & 1 & \texttt{prep} \\

\texttt{Wavelet}
  & Peak count and inter-peak interval statistics across 9 DWT levels
  & 45 & \texttt{prep} \\

\midrule
\textbf{Common features per channel} & & \textbf{22} & \\
\textbf{Five non-\texttt{prep} channels} & & \textbf{$5\times22=110$} & \\
\textbf{Additional \texttt{prep}-only features} & & \textbf{59} & \\
\textbf{Total \texttt{prep} features} & & \textbf{$22+59=81$} & \\
\textbf{Total feature vector} & & \textbf{$110+81=191$} & \\
\bottomrule
\end{tabular}
}
\end{table}

\paragraph{General statistical features} For a windowed signal $x[n]$, $n=1,\ldots,N$, the \texttt{StandardStat} group contains the mean, median, standard deviation, variance, skewness, kurtosis, and range~\cite{joanes1998comparing}. The mean and standard deviation are defined as 
\begin{equation}   
\mu_x =   \frac{1}{N}\sum_{n=1}^{N}x[n],   \qquad   \sigma_x =   \sqrt{   \frac{1}{N-1}\sum_{n=1}^{N}\left(x[n]-\mu_x\right)^2   }.   
\label{eq:mean_std} 
\end{equation} % 

Skewness and kurtosis were computed using standard sample moment definitions~\cite{joanes1998comparing}.

The \texttt{InterQuartile} group contains percentile-based robust statistics. Let $Q_p(x)$ denote the $p$th percentile~\cite{hyndman1996sample} of $x$. The interquartile range and interdecile range are 
\begin{equation}   \mathrm{IQR}   =   Q_{75}(x)-Q_{25}(x),   \qquad   \mathrm{IDR}   =   Q_{90}(x)-Q_{10}(x).   
\label{eq:iqr_idr} \end{equation} 

The individual percentiles $Q_{10}$, $Q_{25}$, $Q_{75}$, and $Q_{90}$ were also retained as features.

\paragraph{Time-domain features}
Let $x[n]$, $n=1,\ldots,N$, denote a discrete BCG signal window, where $N$ is the number of samples in the window, $n$ is the sample index, and $f_s$ is the sampling frequency. The corresponding continuous-time representation is denoted by $x(t)$, with window duration $T=(N-1)/f_s$.

The \texttt{ZeroCrossing}~\cite{rabiner1978digital} group contains the zero-crossing count,
\begin{equation}
  \mathrm{ZC}
  =
  \frac{1}{2}
  \sum_{n=2}^{N}
  \left|
  \operatorname{sign}(x[n])
  -
  \operatorname{sign}(x[n-1])
  \right|,
  \label{eq:zc}
\end{equation}
where $\operatorname{sign}(\cdot)$ returns the sign of its argument. The feature counts sign changes between consecutive samples. It is reported as a count rather than a rate because it is not normalised by the window duration.

The \texttt{AUC} group contains the signed area under the curve, the absolute area under the curve, and the curve length. The signed and absolute AUC features were computed using trapezoidal numerical integration~\cite{press2007numerical}:
\begin{equation}
  \mathrm{AUC}
  =
  \int_{0}^{T} x(t)\,\mathrm{d}t
  \approx
  \frac{1}{f_s}
  \sum_{n=1}^{N-1}
  \frac{x[n]+x[n+1]}{2},
  \label{eq:auc}
\end{equation}
\begin{equation}
  \mathrm{AUC}_{\mathrm{abs}}
  =
  \int_{0}^{T} |x(t)|\,\mathrm{d}t
  \approx
  \frac{1}{f_s}
  \sum_{n=1}^{N-1}
  \frac{|x[n]|+|x[n+1]|}{2}.
  \label{eq:auc_abs}
\end{equation}
Here, $\mathrm{AUC}$ preserves the sign of the signal, whereas $\mathrm{AUC}_{\mathrm{abs}}$ measures the total absolute signal magnitude over the window.

The curve length feature, also referred to as line length in biomedical time-series analysis~\cite{esteller2001line}, was defined as
\begin{equation}
  L_x
  =
  \sum_{n=1}^{N-1}
  \sqrt{
  \left(x[n+1]-x[n]\right)^2
  +
  \left(\frac{1}{f_s}\right)^2
  },
  \label{eq:curve_length}
\end{equation}

where $L_x$ measures the cumulative length of the sampled waveform in the time--amplitude plane.

\paragraph{Frequency-domain features}
Let $x[n]$, $n=1,\ldots,N$, denote the discrete signal window sampled at frequency $f_s$. Its discrete frequency representation~\cite{oppenheim1999discrete} is denoted by $X(f_k)$, where $f_k$ is the frequency associated with FFT bin $k$.

The frequency spacing is
\begin{equation}
  \Delta f = \frac{f_s}{N}.
  \label{eq:frequency_resolution}
\end{equation}

The \texttt{PSD} group was computed using Welch's power spectral density (PSD)~\cite{oppenheim1999discrete} estimate. Let $P(f)$ denote the continuous PSD and $P(f_k)$ its estimate at frequency bin $f_k$. For a frequency interval $[f_1,f_2]$, the corresponding band power was defined as
\begin{equation}
  \mathrm{BP}_{[f_1,f_2]}
  =
  \int_{f_1}^{f_2} P(f)\,\mathrm{d}f
  \approx
  \sum_{f_k \in [f_1,f_2]} P(f_k)\Delta f,
  \label{eq:band_power}
\end{equation}
where $f_1$ and $f_2$ are the lower and upper frequency limits of the band, respectively, and $\Delta f$ is the frequency resolution of the PSD estimate. The retained PSD features were total power over the available frequency range and band power in the intervals 0.1--0.6\,Hz, 0.1--0.4\,Hz, and 0.4--0.8\,Hz.

The \texttt{FFT} group was computed only from the preprocessed (\texttt{prep}) channel. Let $\mathcal{K}$ denote the set of retained FFT bins, restricted to frequencies between 0.05\,Hz and the Nyquist frequency $f_s/2$. For each $k\in\mathcal{K}$, let $f_k$ be the corresponding frequency and
\begin{equation}
  A_k = |X(f_k)|^2
  \label{eq:fft_power}
\end{equation}
the squared FFT magnitude at that frequency.

The dominant frequency was defined as the frequency bin with maximum spectral power:
\begin{equation}
  k^\ast
  =
  \arg\max_{k\in\mathcal{K}} A_k,
  \qquad
  f_{\mathrm{dom}} = f_{k^\ast}.
  \label{eq:domfreq}
\end{equation}

The spectral centroid~\cite{peeters2004large} was defined as the power-weighted mean frequency:
\begin{equation}
  f_c
  =
  \frac{
    \sum_{k\in\mathcal{K}} f_k A_k
  }{
    \sum_{k\in\mathcal{K}} A_k
  },
  \label{eq:spectral_centroid}
\end{equation}
where $f_c$ represents the centre of mass of the spectrum.

For spectral entropy~\cite{peeters2004large}, the spectral powers were first normalised as
\begin{equation}
  p_k
  =
  \frac{A_k}{
    \sum_{j\in\mathcal{K}} A_j
  },
  \qquad
  \sum_{k\in\mathcal{K}}p_k=1.
  \label{eq:normalised_spectral_power}
\end{equation}
The spectral entropy~\cite{peeters2004large} was then computed as
\begin{equation}
  H_{\mathrm{spec}}
  =
  -\sum_{k\in\mathcal{K}} p_k\log p_k.
  \label{eq:spectral_entropy}
\end{equation}
Low spectral entropy indicates that the signal power is concentrated in a small number of frequency bins, whereas high spectral entropy indicates a more dispersed spectrum.

Spectral skewness was computed from the distribution of squared FFT magnitudes $\{A_k\}_{k\in\mathcal{K}}$ and quantifies asymmetry in the spectral-power distribution.

\paragraph{Event-energy features}
The \texttt{Energy} group was computed only from the \texttt{prep} channel. Let $x[n]$, $n=1,\ldots,N$, denote a signal window containing $N$ samples. The total energy, mean power, and root-mean-square~(RMS) amplitude~\cite{oppenheim1999discrete} were defined as
\begin{equation}
  E_{\mathrm{tot}}
  =
  \sum_{n=1}^{N} x[n]^2,
  \qquad
  P_{\mathrm{mean}}
  =
  \frac{1}{N}\sum_{n=1}^{N}x[n]^2,
  \qquad
  \mathrm{RMS}
  =
  \sqrt{P_{\mathrm{mean}}},
  \label{eq:energy}
\end{equation}
where $E_{\mathrm{tot}}$ is the total squared signal amplitude over the window, $P_{\mathrm{mean}}$ is the average squared amplitude per sample, and $\mathrm{RMS}$ is the square root of the mean power.

To describe short-term changes in signal energy, each window was divided into $M$ non-overlapping frames of 1\,s duration. At sampling frequency $f_s$, each frame contains
\begin{equation}
  L = f_s
  \label{eq:frame_length}
\end{equation}
samples. Let $\mathcal{F}_m$ denote the set of sample indices belonging to frame $m$, where $m=1,\ldots,M$. The energy of frame $m$ was defined as
\begin{equation}
  E_m
  =
  \sum_{n\in\mathcal{F}_m}x[n]^2.
  \label{eq:frame_energy}
\end{equation}
The mean, standard deviation, median, and maximum of the frame-energy sequence $\{E_m\}_{m=1}^{M}$ were retained as features. %Any incomplete frame at the end of the signal window was discarded.

\paragraph{Wavelet-derived features}
The \texttt{Wavelet} group was computed only from the \texttt{prep} channel.
A discrete wavelet transform~(DWT) using the biorthogonal \texttt{bior3.9} wavelet~\cite{mallat1989theory, lee2019pywavelets} was applied at nine decomposition levels~\cite{cohen1992biorthogonal}. Let $a^{(\ell)}[r]$ denote the approximation coefficient at coefficient index $r$ and decomposition level $\ell$, where $\ell=1,\ldots,9$.

Peaks were detected separately in the approximation-coefficient sequence at each level. Let
\begin{equation}
  p^{(\ell)}_1,\ldots,p^{(\ell)}_{M_\ell}
\end{equation}
denote the ordered coefficient indices of the $M_\ell$ detected peaks at level $\ell$. The peak count $M_\ell$ was retained as one feature for each decomposition level.

% %
% \begin{equation}
%   M_\ell
%   \label{eq:wavelet_peak_count}
% \end{equation}
% %

When at least two peaks were detected, the inter-peak differences were defined
as
\begin{equation}
  d^{(\ell)}_j
  =
  p^{(\ell)}_{j+1}-p^{(\ell)}_j,
  \qquad
  j=1,\ldots,M_\ell-1.
  \label{eq:wavelet_interpeak}
\end{equation}
Here, $d^{(\ell)}_j$ is the distance, measured in approximation-coefficient indices, between two consecutive peaks at level $\ell$. For each level, the minimum, mean, maximum, and variance of $\{d^{(\ell)}_j\}_{j=1}^{M_\ell-1}$ were retained. Together with the peak count, this gives five features per level and $9\times5=45$ wavelet-derived features in total.

\paragraph{Nonlinear and complexity features}
The \texttt{Entropy} group contains Shannon entropy~\cite{shannon1948mathematical} computed from a histogram of the signal values. Let the range of $x[n]$ be divided into $K$ histogram bins, and let $c_k$ denote the number of samples assigned to bin $k$. The empirical probability of bin $k$ was defined as
\begin{equation}
  p_k
  =
  \frac{c_k}{\sum_{j=1}^{K}c_j},
  \qquad
  \sum_{k=1}^{K}p_k=1.
  \label{eq:histogram_probability}
\end{equation}
The Shannon entropy was then computed as
\begin{equation}
  H
  =
  -\sum_{k=1}^{K}p_k\log p_k,
  \label{eq:shannon_entropy}
\end{equation}
where $H$ quantifies the uncertainty of the signal-value histogram. 
%Histogram bins with $p_k=0$ do not contribute to the sum. 
%In the implementation, $K=64$ bins were used and the natural logarithm was applied.

%~\cite{higuchi1988approach}
The \texttt{HiguchiFD} group contains the Higuchi fractal dimension~\cite{higuchi1988approach}. Let $x[n]$, $n=1,\ldots,N$, denote the signal window, where $N$ is the number of samples. For a scale parameter $k$ and a starting index $m$, the subsampled sequence is
\begin{equation}
  x[m],\,
  x[m+k],\,
  x[m+2k],\,
  \ldots,\,
  x\!\left[m+
  \left\lfloor\frac{N-m}{k}\right\rfloor k\right],
\end{equation}
where $k=1,\ldots,k_{\max}$ and $m=1,\ldots,k$. The length of this subsampled curve was estimated as
\begin{equation}
  L_m(k)
  =
  \frac{N-1}
       {
       \left\lfloor\frac{N-m}{k}\right\rfloor k^2
       }
  \sum_{j=1}^{\left\lfloor\frac{N-m}{k}\right\rfloor}
  \left|
  x[m+jk]-x[m+(j-1)k]
  \right|,
  \label{eq:higuchi_lmk}
\end{equation}
where $L_m(k)$ is the estimated curve length for scale $k$ and starting index $m$. The average curve length at scale $k$ was then defined as
\begin{equation}
  L(k)
  =
  \frac{1}{k}
  \sum_{m=1}^{k}L_m(k).
  \label{eq:higuchi_lk}
\end{equation}
The Higuchi fractal dimension was estimated as the slope of the least-squares linear fit between $\log L(k)$ and $\log(1/k)$ over $k=1,\ldots,k_{\max}$. 
%In the implementation, $k_{\max}=5$.

\subsection{Classification and Evaluation} \label{subsec:classification}

The classification task was binary discrimination between expert-scored respiratory-event windows and event-free windows. The positive class was defined as $y=1$ for respiratory-event windows and the negative class as $y=0$ for event-free windows.

Analyses were implemented in Python using the following packages: \path{numpy}~\cite{oliphant2006guide}, \path{pandas}~\cite{mckinney2011pandas}, \path{scipy}~\cite{virtanen2020scipy} for signal filtering and spectral estimation, and \path{scikit-learn}~\cite{bisong2019introduction} for classification and cross-validation. All models were evaluated using leave-one-patient-out~(LOPO) cross-validation. In each fold, all windows from one participant were held out for testing, and all windows from the remaining participants were used for training. This protocol evaluates whether the extracted BCG features generalise to unseen patients.

Four classifiers were evaluated:

\begin{itemize}   \item \textbf{Majority-class Dummy classifier}: a baseline model that always   predicts the majority class in the training fold.

  \item \textbf{Logistic Regression}: an L2-regularised linear classifier   trained with the \texttt{lbfgs} solver.

  \item \textbf{Random Forest}: a nonlinear ensemble of decision trees trained with class-balanced weighting. Random Forest was also used for feature importance analysis using mean decrease in impurity~(MDI).

  \item \textbf{Histogram Gradient Boosting}: a tree-based gradient boosting   classifier for tabular data. \end{itemize}

Predictive performance was used to verify whether the extracted BCG features contain patient-generalising information about respiratory events. Feature-importance analysis was then used to rank the contribution of the implemented feature groups, individual features, and signal channels. The primary feature-ranking analysis used Random Forest MDI importances. For each trained Random Forest, feature importances were obtained by averaging impurity decrease across trees and normalising the resulting importance values to sum to one.

For each LOPO fold, the following metrics were computed: area under the receiver-operating-characteristic curve~(AUC-ROC), area under the precision-recall curve~(AUC-PR; average precision), F1 score, balanced accuracy, sensitivity~(recall), and specificity. Fold-level metrics were reported as mean~$\pm$~standard deviation and median across held-out patients.

\begin{figure}[ht]   
\centering   
\includegraphics[width=\textwidth]{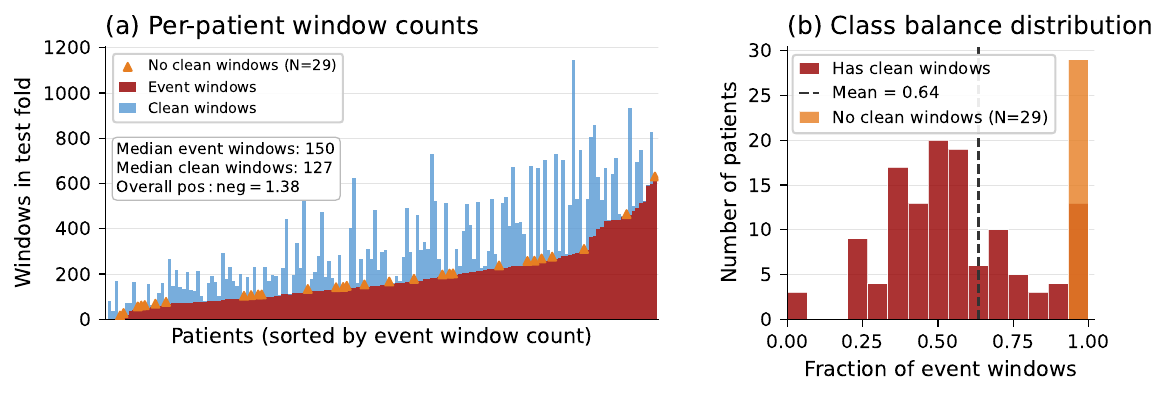}   
\caption{Per-patient class distribution in the held-out test fold.
  \textbf{(a)}~Event~(red) and clean~(blue) windows per patient,
  sorted by event-window count. Orange triangles mark the 29 patients
  whose test fold contained no clean windows after the 30\,s
  event-exclusion buffer; these folds have undefined AUC-ROC but
  well-defined F1 scores.
  \textbf{(b)}~Distribution of the event-window fraction per patient.
  The majority of patients have more event than clean windows~(mean
  fraction $= 0.64$), motivating the use of a balanced class weighting
  in the Random Forest and Hist-GBT.}
\label{fig:balance} 
\end{figure} 
%% ============================================================================
\section{Results}
\label{sec:results}
%% ============================================================================

The final dataset comprised 155 participants~(Table~\ref{tab:demographics}) spanning the full clinical range of OSA severity.

\begin{table}[ht]
\caption{Cohort demographics and dataset statistics. $\pm$ denotes standard deviation.}
\label{tab:demographics}
\centering
\begin{tabular}{lc}
\toprule
\textbf{Characteristic} & \textbf{Value} \\
\midrule
\multicolumn{2}{l}{\textit{Participants}} \\
%Total participants, $N$                         & 155 \\
Sex (female / male)                             & 52 / 103 \\
Age, mean $\pm$ SD (years)                      & 49.5 $\pm$ 12.5 \\
BMI, mean $\pm$ SD (kg/m$^2$)                  & 30.3 $\pm$ 6.9 \\
AHI, median (IQR) (events/h)                   & 14.4 (7.8--27.6) \\
OSA severity (none / mild / moderate / severe)  & 15\% / 36\% / 27\% / 22\% \\
ODI, median (IQR) (events/h)                   & 13.8 (7.5--26.6) \\
SpO$_2$, mean $\pm$ SD (\%)                    & 91.6 $\pm$ 7.9 \\
T90, median (IQR) (\%)                         & 4.2 (0.5--18.7) \\
\midrule
\multicolumn{2}{l}{\textit{Dataset windows (variable duration)}} \\
Event windows (positive class, $y=1$)           & 28,786 \\
Clean windows (negative class, $y=0$)           & 21,088 \\
Positive-to-negative ratio                      & 1.37 \\
Avg.\ event windows per patient                 & 186 $\pm$ 130 \\
\bottomrule
\end{tabular}
\end{table}

\subsection{Synchronisation Quality}
\label{subsec:sync_results}

\begin{figure}[ht] \centering \includegraphics[width=\textwidth]{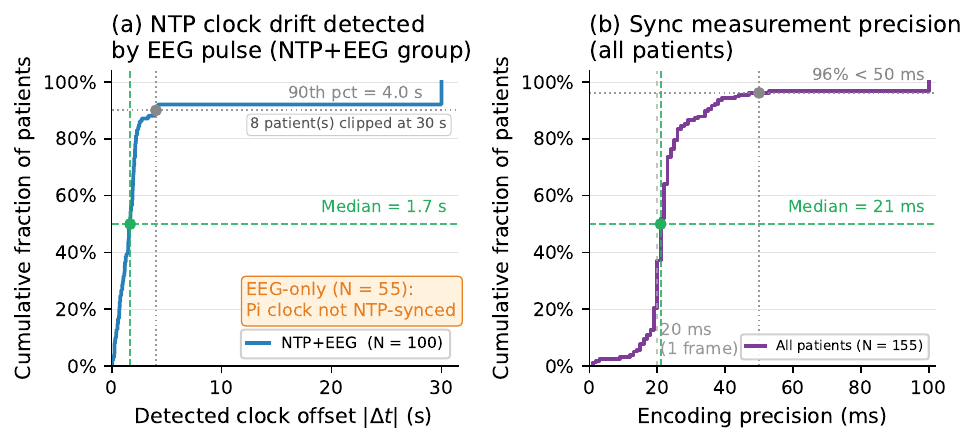} \caption{EEG sync-pulse protocol performance across 155 participants. \textbf{(a)}~Empirical cumulative distribution of the detected absolute clock offset $|\Delta t|$ for the NTP+EEG group. The green dashed line marks the median offset~(1.7\,s), and the grey dotted line marks the 90th percentile~(4.0\,s). Eight participants with offsets above 30\,s are clipped at 30\,s for display. \textbf{(b)}~Encoding quantisation error for all participants~($N=155$), 
defined as the absolute difference between the reconstructed timestamp 
$t_{\mathrm{Pi},0}+\hat{s}$ and the direct wall-clock timestamp logged 
by the RPi at pulse transmission. Median 21\,ms; 96\% of participants 
below 50\,ms.} \label{fig:sync_quality} 
\end{figure}

The EEG sync-pulse protocol was decoded for all 155 recorded participants (Figure~\ref{fig:sync_quality}). Of these, 100 recordings had both NTP-based system-clock information and EEG-pulse synchronisation, while 55 recordings used EEG-pulse synchronisation alone because NTP information was not available. For the NTP+EEG group, the detected absolute clock offset $|\Delta t|$ shows a median of 1.7\,s, with 90\% of participants below 4.0\,s (Figure~\ref{fig:sync_quality}a). These offsets quantify the mismatch between the independent polygraphy and RPi clocks before correction. All detected offsets were corrected before respiratory-event mapping by adding the synchronisation offset $\Delta t$ to each polygraphy event timestamp and converting the corrected time to a BCG sample index, as defined in Eqs.~\ref{eq:event_time_mapping}--\ref{eq:event_sample_mapping}. Each synchronisation pulse encodes a 16-bit integer payload $\hat{s}$, representing the number of whole seconds elapsed since $t_{\mathrm{Pi},0}$, the RPi timestamp of the first BCG sample (as defined in Eq.~\ref{eq:event_sample_mapping}). The reconstructed RPi timestamp corresponding to each decoded frame is therefore $t_{\mathrm{Pi},0} + \hat{s}$. This integer-second encoding introduces a quantisation error that upper-bounds the residual timing uncertainty after synchronisation. To characterise this error, we compared the reconstructed timestamp $t_{\mathrm{Pi},0} + \hat{s}$ to the direct wall-clock timestamp logged by the RPi at the moment of pulse transmission. Across all 155 participants, the median encoding error was 21\,ms (Figure~\ref{fig:sync_quality}b), and 96\% of participants had an error below 50\,ms, confirming that the integer-second approximation introduces negligible residual uncertainty relative to the 30\,s classification window.

\subsection{Classifier Performance}
\label{subsec:clf_results}

Table~\ref{tab:results} summarises LOPO cross-validation performance across the four classifiers. The non-linear ensemble models substantially outperformed both the majority-class baseline and the linear baseline. Random Forest achieved an AUC-ROC of $0.967$ and AUC-PR of $0.977$; Hist-GBT achieved an AUC-ROC of $0.969$ and AUC-PR of $0.979$. Logistic Regression achieved lower performance~(AUC-ROC $=0.792$), indicating that respiratory event discrimination in the BCG feature space is not purely linear. %The high AUC-ROC and AUC-PR of the two non-linear models confirm that the 191-dimensional feature representation contains robust patient-independent information for separating respiratory-event windows from clean windows.

\begin{table}[ht]
\caption{LOPO cross-validation performance: mean $\pm$ SD across $N=155$
  patient folds (AUC-ROC and AUC-PR computed over the 123 folds with
  evaluable negative windows; F1 over all 155). Best values per metric
  are highlighted in \textbf{bold}.}
\label{tab:results}
\centering
\small
\scalebox{0.7}{
\begin{tabular}{lcccccc}
\toprule
\textbf{Classifier} & \textbf{AUC-ROC} & \textbf{AUC-PR} & \textbf{F1}
  & \textbf{Bal.\ Acc.} & \textbf{Sensitivity} & \textbf{Specificity} \\
\midrule
Majority-class
  & $0.500 \pm 0.000$ & $0.565 \pm 0.212$ & $0.742 \pm 0.217$
  & $0.584 \pm 0.212$ & $1.000 \pm 0.000$ & $0.000 \pm 0.000$ \\
Logistic Reg.
  & $0.792 \pm 0.104$ & $0.840 \pm 0.119$ & $0.702 \pm 0.210$
  & $0.705 \pm 0.125$ & $0.678 \pm 0.206$ & $0.736 \pm 0.229$ \\
Random Forest
  & $0.967 \pm 0.044$ & $0.977 \pm 0.036$ & $0.905 \pm 0.153$
  & $0.895 \pm 0.119$ & $\mathbf{0.953 \pm 0.065}$ & $0.807 \pm 0.245$ \\
Hist-GBT
  & $\mathbf{0.969 \pm 0.052}$ & $\mathbf{0.979 \pm 0.040}$
  & $\mathbf{0.915 \pm 0.151}$ & $\mathbf{0.912 \pm 0.100}$
  & $0.944 \pm 0.077$ & $\mathbf{0.862 \pm 0.193}$ \\
\bottomrule
\end{tabular}
}
\end{table}

\begin{figure}[ht]
  \centering
  \includegraphics[width=\textwidth]{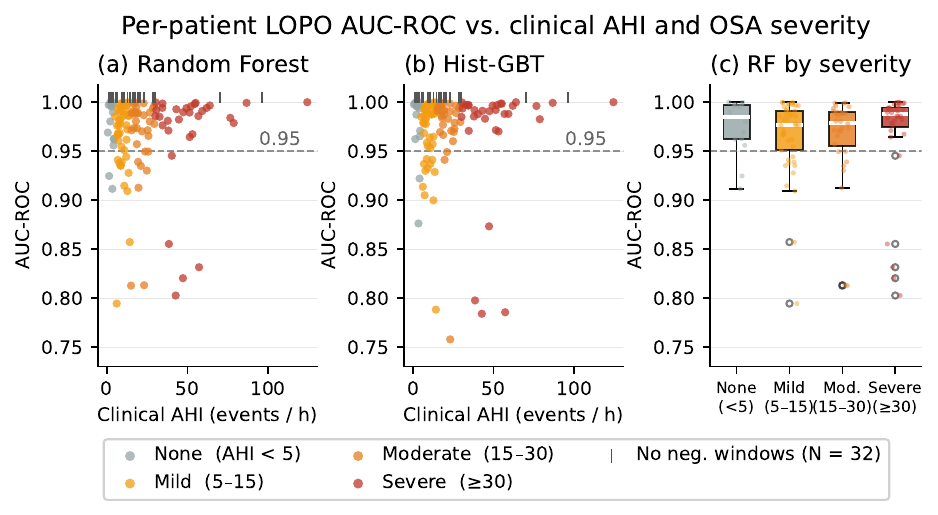}
  \caption{Per-patient LOPO AUC-ROC as a function of clinical AHI and OSA severity, including no OSA, mild OSA, moderate OSA, and severe OSA. \textbf{(a)}~Random Forest and \textbf{(b)}~Hist-GBT: each point represents one held-out patient, coloured by OSA severity category  (AASM thresholds). The dashed horizontal line marks AUC-ROC $= 0.95$.  Vertical tick marks at the top edge indicate the 29 patients whose test  fold contained no eligible negative windows after applying the 30\,s  event-exclusion buffer; AUC-ROC is undefined for these folds~(F1 scores  ranged from 0.90 to 1.00).  \textbf{(c)}~Distribution of Random Forest AUC-ROC stratified by  severity category~(box: IQR; white line: median; whiskers:  $1.5\times$IQR; jittered points: individual folds).  Harder cases~(AUC-ROC $<0.95$) are distributed across all severity  groups, with mild-OSA patients most frequently represented.}  \label{fig:perf_ahi}
\end{figure}

Per-fold AUC-ROC profiles were available for 123 of 155 patients whose test fold contained both respiratory-event and event-free windows. The remaining folds did not contain both classes in the held-out test set, making AUC-ROC undefined for those folds. Among the 123 evaluable folds, 79\% of Random Forest folds and 83\% of Hist-GBT folds yielded AUC-ROC $>0.95$. In addition, 93\% of folds exceeded AUC-ROC $>0.90$ for both models (Figure~\ref{fig:perf_ahi}). Harder folds, defined here as folds with AUC-ROC $<0.95$, were observed across the AHI-defined OSA severity groups used in this study, including mild, moderate, and severe OSA. This indicates that fold difficulty was not determined by AHI severity alone. 
% This supports the interpretation that patient-specific biomechanics, sensor coupling, posture, and respiratory-effort transmission to the mat also influence feature discriminability.

\begin{figure}[ht]
  \centering
  \includegraphics[width=\textwidth]{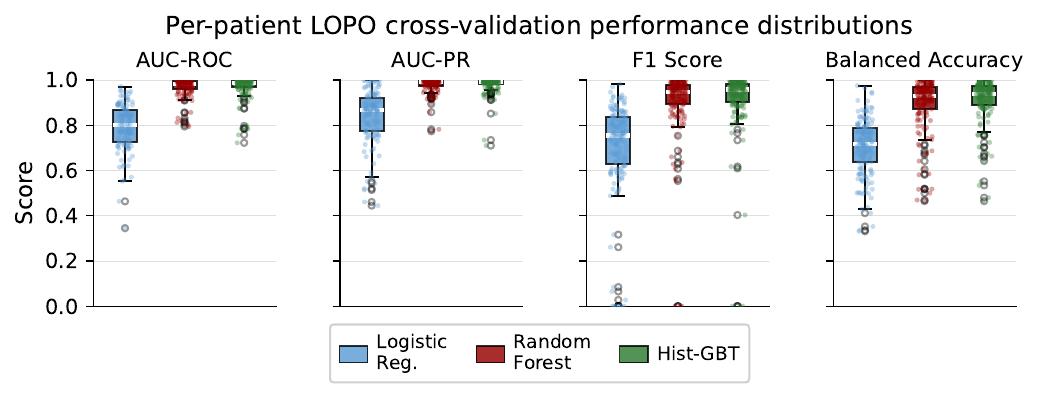}
  \caption{Per-patient LOPO performance distributions shown as
    box-and-whisker plots overlaid with individual fold scores~(jittered
    points) for Logistic Regression, Random Forest, and Hist-GBT across four
    metrics. Boxes span the interquartile range; horizontal white lines show
    medians; whiskers extend to $1.5\times$IQR. The Majority-class Dummy
    baseline is omitted for clarity~(fixed AUC-ROC $=0.5$). Both non-linear
    classifiers exhibit tight, high-valued distributions, confirming robust
    performance across diverse patients.}
  \label{fig:boxplots}
\end{figure}

Figure~\ref{fig:boxplots} summarises the distribution of per-patient fold scores. Both Random Forest and Hist-GBT show compact, high-valued distributions, with AUC-ROC interquartile ranges of 0.961--0.994 and 0.968--0.996, respectively, and medians of 0.982 and 0.988. In contrast, Logistic Regression has a wider AUC-ROC distribution~(IQR 0.728--0.866) and a lower minimum in the hardest folds. This contrast indicates that the implemented feature groups are most effectively exploited by non-linear classifiers, consistent with interactions between signal channel, feature group, patient morphology, and event physiology.

\begin{figure}[ht]
  \centering
  \includegraphics[width=\textwidth]{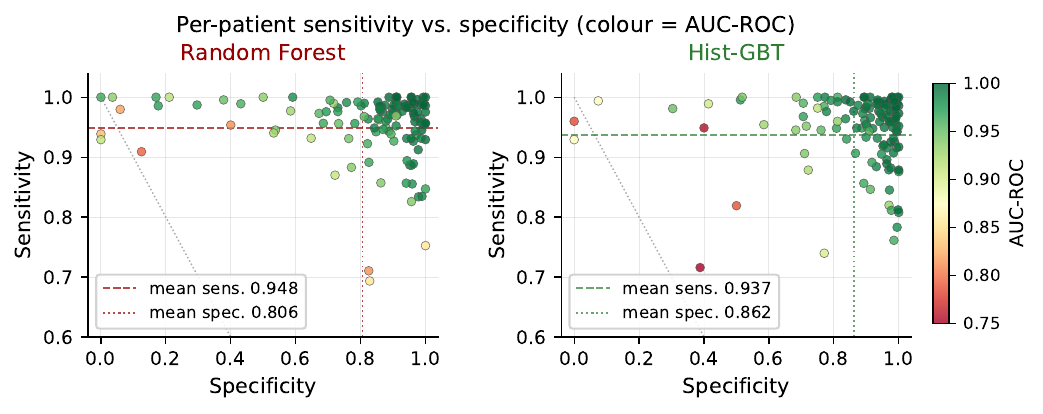}
  \caption{Per-patient sensitivity versus specificity for Random
    Forest~(left) and Hist-GBT~(right), with point colour encoding
    AUC-ROC~(green~=~high, red~=~low). Dashed and dotted lines indicate
    per-classifier mean sensitivity and specificity, respectively. Random
    Forest achieves slightly higher mean sensitivity~($\mu=0.948$) at the
    cost of lower mean specificity~($\mu=0.806$), while Hist-GBT achieves
    a more balanced trade-off~(sensitivity $0.937$, specificity $0.862$),
    making the two classifiers suitable for different clinical deployment
    priorities.}
  \label{fig:sensspec}
\end{figure}

Figure~\ref{fig:sensspec} shows the sensitivity-specificity trade-off at the per-patient level. Most folds lie in the high-sensitivity, high-specificity region. Random Forest is slightly biased toward sensitivity, whereas Hist-GBT provides a more balanced sensitivity-specificity trade-off. For the present study, these differences are secondary to the main feature-discrimination question, but they provide context for possible deployment. %A high-sensitivity operating point may be preferable for closed-loop intervention systems, whereas a more balanced operating point may be preferable for screening scenarios in which false positives increase clinical workload. The few low-specificity outliers are mainly associated with folds containing very few clean control windows; in such cases, one or two false positives can disproportionately reduce specificity.

\subsection{Feature Group Analysis}
\label{subsec:importance}

\begin{figure}[ht]
  \centering
  \includegraphics[width=0.95\textwidth]{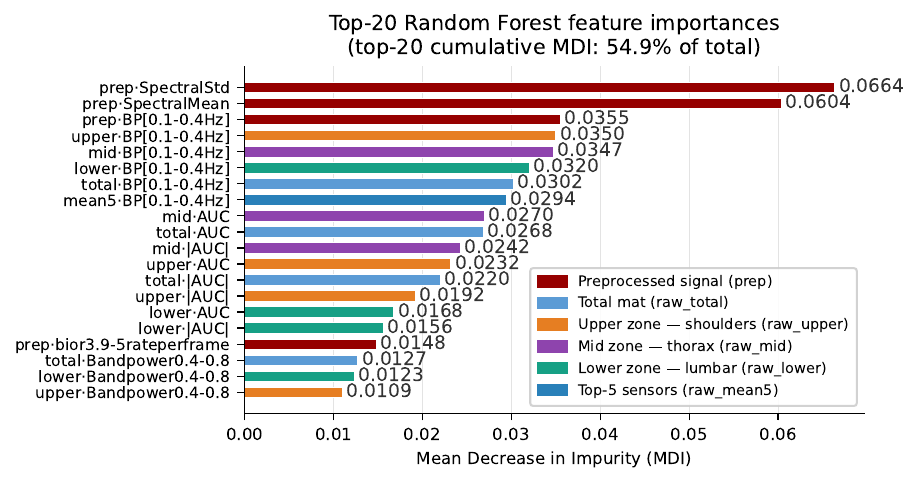}
  \caption{Top-20 Random Forest feature importances ranked by mean decrease
    in impurity~(MDI). Colour encodes the signal channel:
    \textcolor{red}{preprocessed signal (\texttt{prep})},
    \textcolor{blue}{total mat (\texttt{raw\_total})},
    upper zone~(orange, \texttt{raw\_upper}),
    mid zone~(purple, \texttt{raw\_mid}),
    lower zone~(teal, \texttt{raw\_lower}), and
    top-5 sensor mean~(dark blue, \texttt{raw\_mean5}).
    The top-20 features account for 54.9\% of the total MDI. FFT-derived
    frequency descriptors and breathing-band power of \texttt{prep} dominate
    ranks 1--3, with band-power features across all remaining signal channels
    occupying ranks 4--8, and AUC and wavelet-derived features providing
    complementary information.}
  \label{fig:importance}
\end{figure}

Figure~\ref{fig:importance} shows the top-20 Random Forest features ranked by mean decrease in impurity~(MDI). The two highest-ranked features are \path{prep-SpectralStd} and \path{prep-SpectralMean}, both from the \path{FFT} group computed on the preprocessed (\path{prep}) channel. They account for 6.6\% and 6.0\% of the total impurity decrease, respectively. The third-ranked feature is \path{prep-Bandpower0.1-0.4}, a \path{PSD} breathing-band power feature. Ranks 4--8 are occupied by the same $0.1$--$0.4$\,Hz band-power feature across the remaining raw signal channels~(\path{raw_upper}, \path{raw_mid}, \path{raw_lower}, \path{raw_total}, and \path{raw_mean5}). Together, the top-20 features account for 54.9\% of the total MDI.

% This hierarchy indicates that respiratory-event discrimination is dominated by two related but distinct sources of information. First, the \texttt{prep} channel concentrates respiratory dynamics sufficiently for FFT-derived frequency descriptors to become highly discriminative. Second, breathing-band power in the $0.1$--$0.4$\,Hz range remains informative across all signal channels, indicating that the event-related respiratory signature is distributed over the BCG rather than confined to a single region. The presence of AUC features and wavelet-derived features among the top-20 suggests that time-domain integral features and multi-scale temporal descriptors add complementary information, but the dominant feature tier is frequency-domain.

Figure~\ref{fig:importance} reports the leading Random Forest feature
importances used in the main analysis. To support a more detailed
interpretation of the discriminative signal structure,
Table~\ref{tab:app_top30_importance}, Table~\ref{tab:app_group_importance},
and Table~\ref{tab:app_channel_importance} report the top-30 individual
features and the corresponding summaries aggregated by implemented feature
group and signal channel. Importances are Random Forest MDI values computed
on the full feature set and normalised to sum to one across all 191 features.

\begin{table*}[ht]
\caption{Top-30 Random Forest feature importances. Importance denotes
normalised mean decrease in impurity~(MDI). The cumulative column reports
the cumulative percentage of total MDI explained up to each rank.}
\label{tab:app_top30_importance}
\centering
\scriptsize
\setlength{\tabcolsep}{3.0pt}
\resizebox{\textwidth}{!}{%
\begin{tabular}{rlllr r}
\toprule
\textbf{Rank} & \textbf{Feature} & \textbf{Signal channel} &
\textbf{Feature group} & \textbf{Importance} & \textbf{Cumulative} \\
\midrule
1 & \texttt{prep-SpectralStd} & \texttt{prep} & \texttt{FFT} & 0.0664 & 6.6\% \\
2 & \texttt{prep-SpectralMean} & \texttt{prep} & \texttt{FFT} & 0.0604 & 12.7\% \\
3 & \texttt{prep-Bandpower0.1-0.4} & \texttt{prep} & \texttt{PSD} & 0.0355 & 16.2\% \\
4 & \texttt{raw\_upper-Bandpower0.1-0.4} & \texttt{raw\_upper} & \texttt{PSD} & 0.0350 & 19.7\% \\
5 & \texttt{raw\_mid-Bandpower0.1-0.4} & \texttt{raw\_mid} & \texttt{PSD} & 0.0347 & 23.2\% \\
6 & \texttt{raw\_lower-Bandpower0.1-0.4} & \texttt{raw\_lower} & \texttt{PSD} & 0.0320 & 26.5\% \\
7 & \texttt{raw\_total-Bandpower0.1-0.4} & \texttt{raw\_total} & \texttt{PSD} & 0.0302 & 29.5\% \\
8 & \texttt{raw\_mean5-Bandpower0.1-0.4} & \texttt{raw\_mean5} & \texttt{PSD} & 0.0294 & 32.4\% \\
9 & \texttt{raw\_mid-AreaUnderCurve} & \texttt{raw\_mid} & \texttt{AUC} & 0.0270 & 35.1\% \\
10 & \texttt{raw\_total-AreaUnderCurve} & \texttt{raw\_total} & \texttt{AUC} & 0.0268 & 37.8\% \\
11 & \texttt{raw\_mid-AreaUnderCurveAbs} & \texttt{raw\_mid} & \texttt{AUC} & 0.0242 & 40.2\% \\
12 & \texttt{raw\_upper-AreaUnderCurve} & \texttt{raw\_upper} & \texttt{AUC} & 0.0232 & 42.5\% \\
13 & \texttt{raw\_total-AreaUnderCurveAbs} & \texttt{raw\_total} & \texttt{AUC} & 0.0220 & 44.7\% \\
14 & \texttt{raw\_upper-AreaUnderCurveAbs} & \texttt{raw\_upper} & \texttt{AUC} & 0.0192 & 46.6\% \\
15 & \texttt{raw\_lower-AreaUnderCurve} & \texttt{raw\_lower} & \texttt{AUC} & 0.0168 & 48.3\% \\
16 & \texttt{raw\_lower-AreaUnderCurveAbs} & \texttt{raw\_lower} & \texttt{AUC} & 0.0156 & 49.9\% \\
17 & \texttt{prep-bior3.9-5rateperframe} & \texttt{prep} & \texttt{Wavelet} & 0.0148 & 51.4\% \\
18 & \texttt{raw\_total-Bandpower0.4-0.8} & \texttt{raw\_total} & \texttt{PSD} & 0.0127 & 52.6\% \\
19 & \texttt{raw\_lower-Bandpower0.4-0.8} & \texttt{raw\_lower} & \texttt{PSD} & 0.0123 & 53.9\% \\
20 & \texttt{raw\_upper-Bandpower0.4-0.8} & \texttt{raw\_upper} & \texttt{PSD} & 0.0109 & 54.9\% \\
21 & \texttt{prep-bior3.9-5vardiff} & \texttt{prep} & \texttt{Wavelet} & 0.0106 & 56.0\% \\
22 & \texttt{prep-SpectralSkewness} & \texttt{prep} & \texttt{FFT} & 0.0101 & 57.0\% \\
23 & \texttt{prep-DomFreq} & \texttt{prep} & \texttt{FFT} & 0.0100 & 58.0\% \\
24 & \texttt{raw\_mid-Bandpower0.4-0.8} & \texttt{raw\_mid} & \texttt{PSD} & 0.0099 & 59.0\% \\
25 & \texttt{prep-bior3.9-5meandiff} & \texttt{prep} & \texttt{Wavelet} & 0.0098 & 60.0\% \\
26 & \texttt{raw\_upper-Bandpower0.1-0.6} & \texttt{raw\_upper} & \texttt{PSD} & 0.0097 & 60.9\% \\
27 & \texttt{prep-bior3.9-6meandiff} & \texttt{prep} & \texttt{Wavelet} & 0.0082 & 61.8\% \\
28 & \texttt{prep-Bandpower0.4-0.8} & \texttt{prep} & \texttt{PSD} & 0.0077 & 62.5\% \\
29 & \texttt{raw\_upper-Mean} & \texttt{raw\_upper} & \texttt{StandardStat} & 0.0076 & 63.3\% \\
30 & \texttt{raw\_total-HiguchiFD} & \texttt{raw\_total} & \texttt{HiguchiFD} & 0.0075 & 64.0\% \\
\bottomrule
\end{tabular}
}
\end{table*}

The top-30 features account for 64.0\% of the total MDI. The leading features
are dominated by \texttt{PSD} breathing-band power, \texttt{FFT}-derived
frequency descriptors, and \texttt{AUC}-based time-domain integral features.
The highest-ranked \texttt{Wavelet} feature is the peak-count feature from
the \texttt{bior3.9} decomposition of \texttt{prep}. This might indicate that respiratory-event discrimination is driven primarily by
disruption of breathing-related spectral structure, with complementary
information from integrated pressure displacement and multi-scale temporal
structure.

\begin{table}[ht]
\caption{Random Forest importance aggregated by implemented feature group.
Values are normalised MDI summed across all features in each group.}
\label{tab:app_group_importance}
\centering
\small
\begin{tabular}{lrr}
\toprule
\textbf{Feature group} & \textbf{Importance} & \textbf{Share of total} \\
\midrule
\texttt{PSD}              & 0.3029 & 30.3\% \\
\texttt{AUC}              & 0.2154 & 21.5\% \\
\texttt{FFT}              & 0.1508 & 15.1\% \\
\texttt{InterQuartile}    & 0.1206 & 12.1\% \\
\texttt{StandardStat}     & 0.1048 & 10.5\% \\
\texttt{Wavelet}          & 0.0687 & 6.9\% \\
\texttt{HiguchiFD}        & 0.0239 & 2.4\% \\
\texttt{Energy}           & 0.0094 & 0.9\% \\
\texttt{ZeroCrossing}     & 0.0019 & 0.2\% \\
\texttt{Entropy}          & 0.0015 & 0.2\% \\
\bottomrule
\end{tabular}
\end{table}

Aggregating importance by feature group (Table \ref{tab:app_channel_importance})
 confirms the pattern seen in the
individual ranking. \texttt{PSD} features explain the largest share of total
importance~(30.3\%), followed by \texttt{AUC} features~(21.5\%) and
\texttt{FFT}-derived frequency descriptors~(15.1\%). Together, these three
groups account for 66.9\% of the total MDI. \texttt{Wavelet}-derived
features contribute 6.9\%, while \texttt{HiguchiFD}, \texttt{Energy},
\texttt{ZeroCrossing}, and \texttt{Entropy} features contribute smaller
shares under the present sensor, model and analysis.

\begin{table}[ht]
\caption{Random Forest importance aggregated by signal channel. Values are
normalised MDI summed across all features extracted from each channel.}
\label{tab:app_channel_importance}
\centering
\small
\begin{tabular}{lrr}
\toprule
\textbf{Signal channel} & \textbf{Importance} & \textbf{Share of total} \\
\midrule
\texttt{prep}       & 0.3074 & 30.7\% \\
\texttt{raw\_upper} & 0.1656 & 16.6\% \\
\texttt{raw\_total} & 0.1557 & 15.6\% \\
\texttt{raw\_mid}   & 0.1553 & 15.5\% \\
\texttt{raw\_lower} & 0.1420 & 14.2\% \\
\texttt{raw\_mean5} & 0.0741 & 7.4\% \\
\bottomrule
\end{tabular}
\end{table}

Table~\ref{tab:group_ranking} summarises the feature-group ranking and provides a practical recommendation for each group.

\begin{table}[ht]
\caption{Feature-group MDI ranking. MDI\,\% is the fraction of total
  Random Forest mean decrease in impurity attributed to each group,
  averaged across all 155 LOPO folds. Per-feature average~=~MDI\,\%
  divided by the number of features in the group.
  Recommendation: \textbf{Include}~=~high information per feature;
  \textbf{Consider}~=~moderate contribution spread over many features;
  \textbf{Optional}~=~low per-feature contribution.}
\label{tab:group_ranking}
\centering
\scalebox{0.8}{
\begin{tabular}{llrrrc}
\toprule
\textbf{Feature group} &
\textbf{Applied to} &
\textbf{N} &
\textbf{MDI\,\%} &
\textbf{Per-feat.\,\%} &
\textbf{Rec.} \\
\midrule
Freq-domain: PSD band power     & All channels & 24 & 30.3 & 1.26 & \textbf{Include} \\
Time-domain: AUC / curve length & All channels & 18 & 21.5 & 1.20 & \textbf{Include} \\
Freq-domain: FFT shape          & \path{prep} only & 6  & 15.1 & 2.51 & \textbf{Include} \\
Gen.\ stat: interquartile range & All channels & 36 & 12.1 & 0.33 & Consider \\
Gen.\ stat: central/spread      & All channels & 42 & 10.5 & 0.25 & Consider \\
Wavelet (\path{bior3.9})      & \path{prep} only & 45 & 6.9  & 0.15 & Optional \\
Non-linear: Higuchi FD          & All channels &  6 &  2.4 & 0.40 & Optional \\
Frame energy                    & \path{prep} only &  7 &  0.9 & 0.13 & Optional \\
Time-domain: zero-crossing      & All channels &  6 &  0.2 & 0.03 & Optional \\
Non-linear: Shannon entropy     & \path{prep} only &  1 &  0.2 & 0.15 & Optional \\
\midrule
\textbf{Total}                  &              & \textbf{191} & \textbf{100.0} & & \\
\bottomrule
\end{tabular}
}
\end{table}

% The channel-level summary shows that \texttt{prep} is the most informative single representation, accounting for 30.7\% of total MDI. However, the raw global and zonal channels together account for 61.9\% of the importance, indicating that respiratory-event information remains spatially distributed across the mat. This supports the use of both adaptive top-sensor preprocessing and raw spatial aggregates in the current feature representation.

%% ============================================================================ 

%% ============================================================================
\section{Discussion}
\label{sec:discussion}
%% ============================================================================

\subsection{Dominant Role of Frequency-Domain Features}
\label{subsec:disc_freq}

The most prominent finding is that frequency-domain features account for 45.4\% of the total MDI: the FFT spectral-shape sub-group~(6 features, 15.1\%) and the PSD band-power sub-group~(24 features, 30.3\%) together dominate the feature ranking. On a per-feature basis, FFT spectral-shape descriptors average 2.51\% MDI each, the highest average of any group. The two most important individual features, \path{prep-SpectralStd} and \path{prep-SpectralMean}, account for 6.6\% and 6.0\% of total MDI respectively.

This dominance has a direct physiological basis. During normal breathing, the mat records quasi-periodic pressure modulation caused by thoracoabdominal motion, which is the signal component targeted by several BCG systems~\cite{waltisberg2016detecting,gao2019obstructive,qi2023mattress}. During apnea or hypopnea, this modulation is altered, interrupted, or distorted, and the recovery phase may introduce deeper breaths or irregular motion~\cite{berry2012rules,waltisberg2016detecting}. Because the BCG signal in the adult breathing range~($0.1$--$0.4$\,Hz) is driven by this effort, an apneic event produces a characteristic change of power in that band and a concurrent shift in the centroid and spread of the spectral distribution. FFT spectral-shape descriptors capture the full shape of this shift, whereas band-power features quantify its magnitude independently in each spatial channel. The consistent appearance of band-power features across all six spatial channels confirms that the respiratory signature is detectable across the entire mat surface, regardless of whether a global aggregate, a zonal aggregate, or a per-window top-sensor mean, in which the five sensors with the strongest respiratory periodicity are reselected for each analysis window.

\subsection{Information Content of Time-Domain AUC Features}
\label{subsec:disc_td}

AUC and curve-length features~(18 features across six channels) account for 21.5\% of total MDI, making them the second most informative category overall and the main source of information complementary to the frequency-domain group. On a per-feature basis their average contribution~(1.20\%) is comparable to that of band-power features~(1.26\%), indicating that these two groups are similarly efficient, despite capturing different signal properties.

The importance of AUC features is consistent with the physical nature of BCG sensing, where the recorded signal reflects a mixture of respiratory effort, cardiac-related micromotion, posture, and gross body movement~\cite{giovangrandi2011ballistocardiography,waltisberg2016detecting,vitazkova2024advances}. A respiratory event can reduce the regular respiratory component, but it may also change the integrated pressure variation through altered respiratory effort, recovery breaths, arousal-related movement, or posture-dependent pressure redistribution~\cite{waltisberg2016detecting}. Curve length additionally captures the smoothness of the waveform over time: regular breathing produces a quasi-periodic oscillation with a characteristic arc-length per cycle, while apneic interruptions and recovery breaths alter this pattern. Together, these time-domain descriptors provide a complementary representation to spectral measures, because they are sensitive to changes in the amplitude dynamics of the pressure waveform rather than its frequency structure.

\subsection{Limited Contribution of Wavelet and Nonlinear Features} \label{subsec:disc_wave}

The wavelet group~(\path{bior3.9}, 9-level decomposition, 45 features) accounts for 6.9\% of total MDI, averaging only 0.15\% per feature. This is the lowest per-feature average of any group, meaning that a very large number of features provides a small and diffuse contribution. The Higuchi fractal dimension group~(6 features, 2.4\%) and the frame-energy group~(7 features, 0.9\%) make similarly limited contributions. Shannon entropy~(1 feature, 0.2\%) and zero-crossing rate~(6 features, 0.2\%) are the two least informative groups.

These findings do not imply that lower-ranked descriptors are physiologically meaningless~\cite{richman2000physiological, wang2017assessing}. However, under this validation protocol and this feature library, they do not provide information independent of what is already captured by the spectral and time-domain groups. This may reflect the high signal quality achieved by the adaptive sensor selection in the present setup: when the respiratory periodicity is already well concentrated in the averaged channel, wavelet decomposition and nonlinear transforms may offer limited additional discriminability compared with direct spectral measures. In settings with lower signal-to-noise ratio, for example, lower-resolution pressure mats or sensors with weaker body coupling, these transforms could recover information that spectral features miss, and their contribution may therefore differ across hardware configurations. What is clear from the data is that, for a BCG respiratory-event detector using the features studied here, wavelet-derived and nonlinear features are secondary and can be deprioritised without substantial loss of discriminative information.

\subsection{Effect of Preprocessing on Feature Discriminability} \label{subsec:disc_prep}

A consistent pattern across the ranking is that features computed on the \path{prep} channel (the adaptively preprocessed, polarity-aligned spatial average) dominate the top individual ranks. The top three individual features all belong to \path{prep} (\path{prep-SpectralStd}, \path{prep-SpectralMean}, \path{prep-Bandpower0.1-0.4}). This confirms that the adaptive sensor selection and polarity alignment concentrate the respiratory signature into a single averaged signal, amplifying the discriminability of frequency-domain features computed on that channel.

Raw spatial aggregates~(\path{raw_total}, \path{raw_upper}, \path{raw_mid}, \path{raw_lower}, \path{raw_mean5}) also contribute substantially through their band-power features, occupying ranks 4--10 for the 0.1--0.4\,Hz band. This shows that the respiratory band-power reduction is a robust, spatially distributed signature that does not require extensive preprocessing to be detectable, but that preprocessing specifically benefits the spectral-shape features that distinguish normal breathing from apneic quiescence by the shape ``not just the magnitude'' of the power spectrum.

\subsection{Practical Feature-Group Recommendations} \label{subsec:disc_practical}

The ranking in Table~\ref{tab:group_ranking} supports three concrete design recommendations for future BCG systems targeting respiratory-event detection. Frequency-domain and AUC features should be included in any implementation: together they account for 67\% of total discriminative information while requiring only 48 features across six channels, and both can be computed efficiently from short signal segments using standard spectral and integration methods.

General statistical descriptors (central tendency, spread, and interquartile range) account for a further 22.6\% of MDI and are cheap to compute, but they are spread over 78 features, giving a low per-feature average of 0.29\%. They complement the dominant groups through a different representation of amplitude behaviour and should be included when computational budget permits, but they are not essential for acceptable performance.

Wavelet-derived features~(45 features, 6.9\%), Higuchi fractal dimension~(6 features, 2.4\%), frame energy~(7 features, 0.9\%), and Shannon entropy~(1 feature, 0.2\%) together account for only 10.4\% of MDI distributed over 59 features, each contributing on average less than 0.18\% individually. In resource-constrained or real-time implementations, removing these groups would reduce the feature vector by 31\% while sacrificing only 10\% of discriminative information, making their omission a reasonable engineering trade-off in most deployment scenarios.

\subsection{Comparison with Prior Work} \label{subsec:disc_compare}

The AUC-ROC values of 0.967~(Random Forest) and 0.969~(Hist-GBT) under strict LOPO validation compare favourably with prior BCG-based OSA detection studies. However, direct comparison with previous work is limited because studies differ substantially in sensing hardware, feature design, cohort size, prediction target, and validation protocol. BCG studies reporting AUC above 0.9 have generally used smaller cohorts~($N < 50$) and cross-validation schemes that do not fully prevent data leakage across patients~\cite{wang2017assessing,siyahjani2022performance}. In addition, prior studies have often focused on specific subsets of information, such as BCG-derived heartbeat-interval features, HRV-related frequency bands, nonlinear interval dynamics, respiratory waveform similarity, or sensor-fusion strategies, rather than evaluating a common set of time-domain, frequency-domain, wavelet, complexity, and spatial pressure-mat features under the same patient-independent protocol. Waltisberg et al.\ reported F1 scores of 0.715 for breathing irregularity detection with a bed-based BCG system in nine participants~\cite{waltisberg2016detecting}, considerably lower than the 0.905--0.915 F1 reported here. This difference should therefore be interpreted in the context of both cohort size and feature representation. The primary contribution of the present work is not only the absolute performance level, but the systematic feature ranking it provides: the results show that respiratory-event detection benefits from complementary information in adaptive frequency-domain descriptors, multi-channel breathing-band power, and time-domain AUC features. This ranking can inform future BCG and pressure-mat systems by indicating which feature groups are most important to preserve when designing compact real-time pipelines. 

\subsection{Limitations} \label{subsec:disc_limits} 

Several limitations should be acknowledged. First, the reference standard was a cardiorespiratory polygraphy rather than full polysomnography. Consequently, hypopneas were scored only when accompanied by oxygen desaturation, and cortical arousal-based hypopneas were not captured. This may lead to an underestimate of the true apnea-hypopnea index in patients whose events terminate primarily with arousal, and the feature ranking reported here applies specifically to the desaturation-confirmed event definition.

Second, central respiratory events were present in the cohort but represented a small minority of scored events; the classification framework pools all respiratory event types and does not distinguish between obstructive, mixed, and central subtypes. A finer annotation target may yield a different feature ranking.

Third, the feature importance ranking is based on Random Forest MDI, which can overestimate the contribution of features with many unique values and is not a measure of causal importance. Fourth, it should be noted that Waltisberg et al.~\cite{waltisberg2016detecting} targeted breathing irregularity detection rather than individual event classification, and Wang et al.~\cite{wang2017assessing} performed severity classification at the subject level rather than window-level binary detection. These differences in prediction target mean that the performance values are not directly comparable, and the discussion should be read as a comparison of feature sets and validation methodology rather than of classification accuracy. Fifth, the study was conducted in a controlled in-laboratory environment with a standardised mattress and bed height; the extent to which the feature ranking holds in home settings with varying mattress types, body positions, and co-sleeping remains to be established. Sixth, the wavelet features implemented here use a specific \path{bior3.9} basis with inter-peak interval statistics; other wavelet representations may yield different group-level contributions.

%% ============================================================================
\section{Conclusion}
\label{sec:conclusion}
%% ============================================================================

We have reported a systematic, patient-independent comparison of ten feature groups for BCG-based respiratory-event detection in a cohort of 155 patients. Frequency-domain features are the dominant category, accounting for 45.4\% of total Random Forest discriminative information: PSD band power in the breathing range~($0.1$--$0.4$\,Hz) contributes 30.3\% across all six signal channels, and FFT spectral-shape descriptors contribute a further 15.1\% on the preprocessed channel alone. These two groups together outperform all other categories combined, each averaging more than 1.2\% MDI per feature.

Adaptive preprocessing amplifies this discriminability: the top three individual features all belong to the preprocessed channel~(\path{prep}), and FFT spectral-shape features computed on \path{prep} average 2.51\% MDI per feature, approximately twice the per-feature average of the same descriptors applied to raw channels. High-pass filtering, quality-score-based sensor selection, and polarity-aligned averaging concentrate the respiratory signature and specifically benefit spectral-shape features.

AUC and curve-length features are the main complementary group, contributing 21.5\% of MDI through sensitivity to changes in cumulative pressure displacement during events. General statistical features provide a further 22.6\% but are distributed over 78 features~(0.29\% per feature) and are therefore less efficient. Wavelet-derived features, Higuchi fractal dimension, frame energy, and entropy together contribute 10.4\% across 59 features and represent the weakest tier.

The practical implication is direct: a BCG respiratory-event detector can achieve near-maximal discriminative performance using only the frequency-domain and time-domain AUC groups~(48 features, 67\% of MDI). General statistical features provide a modest further gain; wavelet-derived, frame-energy, and nonlinear features can be omitted from resource-constrained implementations without substantial performance loss. This feature ranking provides an empirical reference point for the design and evaluation of future BCG systems for sleep-disordered-breathing detection.
\section*{Acknowledgements}

This research was funded by the Swiss National Science Foundation (SNSF) grant number 32003BM\_220059.

\section*{Declaration of Competing Interest}

The authors declare no competing interests.

%% ============================================================================
%% Bibliography
%% ============================================================================
\bibliographystyle{elsarticle-num}

\bibliography{bibliography}
%% Inline bibliography — replace with \bibliography{refs} if using BibTeX

\end{document}